\documentclass[10pt,twocolumn,letterpaper]{article}
\usepackage[pagenumbers]{cvpr}
\usepackage[T1]{fontenc}
\usepackage{array}
\usepackage{microtype}
\usepackage{stfloats}
\usepackage{placeins}
\usepackage{float}
\newcommand{\pifive}{\ensuremath{\pi_{0.5}}}

\definecolor{cvprblue}{rgb}{0.21,0.49,0.74}
\usepackage[pagebackref,breaklinks,colorlinks,allcolors=cvprblue]{hyperref}
\renewcommand{\paragraph}[1]{\par\addvspace{.5em}\noindent\textbf{#1}\enspace\ignorespaces}
\title{Sight: Anchoring Execution Intent in Robotic Policies}
\author{%
  Yifan Wang\textsuperscript{1,*}\quad
  Yilong Qiu\textsuperscript{2,*}\quad
  Shuo Bao\textsuperscript{3}\quad
  Yuchen Huang\textsuperscript{4}\\
  Han Yu\textsuperscript{2}\quad
  Jintao Xu\textsuperscript{2}\quad
  Minghua Liu\textsuperscript{5}\quad
  Shuo Tian\textsuperscript{6}\\
  Ling Tak Chung\textsuperscript{7}\quad
  Jiangtao Wang\textsuperscript{8}\quad
  Xinzhou Wang\textsuperscript{9,\textdagger}\quad
  Fuchun Sun\textsuperscript{10,\textdagger}\\[0.5em]
  \small \textsuperscript{1}Harbin Institute of Technology\quad
  \textsuperscript{2}Beijing Institute of Technology\quad
  \textsuperscript{3}Peking University\\
  \small \textsuperscript{4}University of Science and Technology Beijing\quad
  \textsuperscript{5}University of the Chinese Academy of Sciences\\
  \small \textsuperscript{6}Beijing University of Chemical Technology\quad
  \textsuperscript{7}University College London\quad
  \textsuperscript{8}Monash University\\
  \small \textsuperscript{9}Tongji University\quad
  \textsuperscript{10}Tsinghua University%
}
\begin{document}
\raggedbottom
\maketitle
\begingroup
\renewcommand{\thefootnote}{\fnsymbol{footnote}}
\footnotetext[1]{Yifan Wang and Yilong Qiu contributed equally.}
\footnotetext[2]{Xinzhou Wang and Fuchun Sun are corresponding authors.}
\endgroup
\begin{abstract}
Task-specific post-training equips robotic policies with complex manipulation skills, yet executing a feasible motion does not ensure acting on the instructed object with the instructed arm. In dual-arm, multi-object scenes, familiar scene--action associations can compete with language and lead to unintended execution choices. We introduce \textbf{Sight}, a post-training framework that grounds execution intent through a common arm--target assignment. Embodied Collaborative Representation expresses the task, executing arm, and distinguishable target in the language condition used by both the policy and the offline region predictor. Interaction-Centric Spatial Anchoring translates this assignment into target-and-end-effector region priors and assignment-conditioned attention supervision: the head and active wrist provide supervision for single-arm operations, and all three views for simultaneous dual-arm operations. Sight retains complete RGB inputs and the original deployment interface. Integrated into \pifive{}, it improves mean success on four RoboTwin 2.0 tasks by 13.75 percentage points (34.0\% relative). With two additional distractor bottles, success increases from 8\% to 16\%. On the Tianji M6 robot, rounded mean success across arm specification, target selection, and joint assignment increases from 47\% to 83\%. Under the same explicit language, Sight exceeds ECR-only post-training by 11.50 points in simulation. Applied to OpenVLA-OFT, Sight raises mean success from 11.25\% to 21.50\%, a gain of 10.25 points.
\end{abstract}

\section{Introduction}

Task-specific post-training allows generalist robotic policies to apply pretrained manipulation capabilities to a broad range of tasks.~\cite{openvla,octo,pi0,pi05} Yet when a scene admits several feasible actions, a policy can produce executable motion while using an unintended arm or acting on the wrong object. In dual-arm, multi-object settings, this mismatch can appear as an arm or object preference, or as successive action chunks approaching different candidate targets. Reliable manipulation therefore requires the policy to bind its capabilities to the \textbf{arm--target relationship specified by the current instruction} and maintain that relationship while the task requires it.

Correlations between scenes and demonstrated actions can obscure this requirement. A policy may reproduce familiar trajectories without learning to change its arm or target selection when the instruction changes. Recent studies show that target information can be encoded in policy representations without controlling the selected object; valid instruction substitutions in an unchanged scene further expose the gap between visual preferences and language conditioning.~\cite{groundinggap,cag,instructmove} Dual-arm manipulation adds a coupled decision: selecting the correct object is insufficient if the wrong arm executes the action. Execution intent thus concerns both the joint assignment of arm and target and its use during continued execution.

The learning problem has two connected parts. Demonstrations must first distinguish the task requirement, executing arm, and target identity in their language conditions. Those conditions must then refer to the corresponding visual instances and end effectors. Explicit descriptions identify the required assignment, while spatial supervision identifies its visual evidence. For dual-arm policies, the assignment also determines which wrist observes the executing end effector. This motivates our design principle: \textbf{use the demonstrated arm--target relationship to specify both what is supervised and which views provide that supervision}. Language, regional labels, and auxiliary observation sources then describe a common execution choice.

We introduce \textbf{Sight}, an execution-intent anchoring framework with two components. \textbf{Embodied Collaborative Representation} (ECR) organizes task requirements, the executing arm, and a distinguishable target into a unified language condition. \textbf{Interaction-Centric Spatial Anchoring} (ICSA) uses that assignment to construct region priors over the target and active end effector and supervise policy attention. The assignment also determines the sources of auxiliary supervision: the head view supplies scene-level evidence, while the active-wrist view supplies evidence local to the executing arm. Region priors are generated offline; training and deployment both retain all three RGB inputs, and inference requires no additional region predictor.

\begin{figure*}[!t]
\centering
\includegraphics[width=\textwidth]{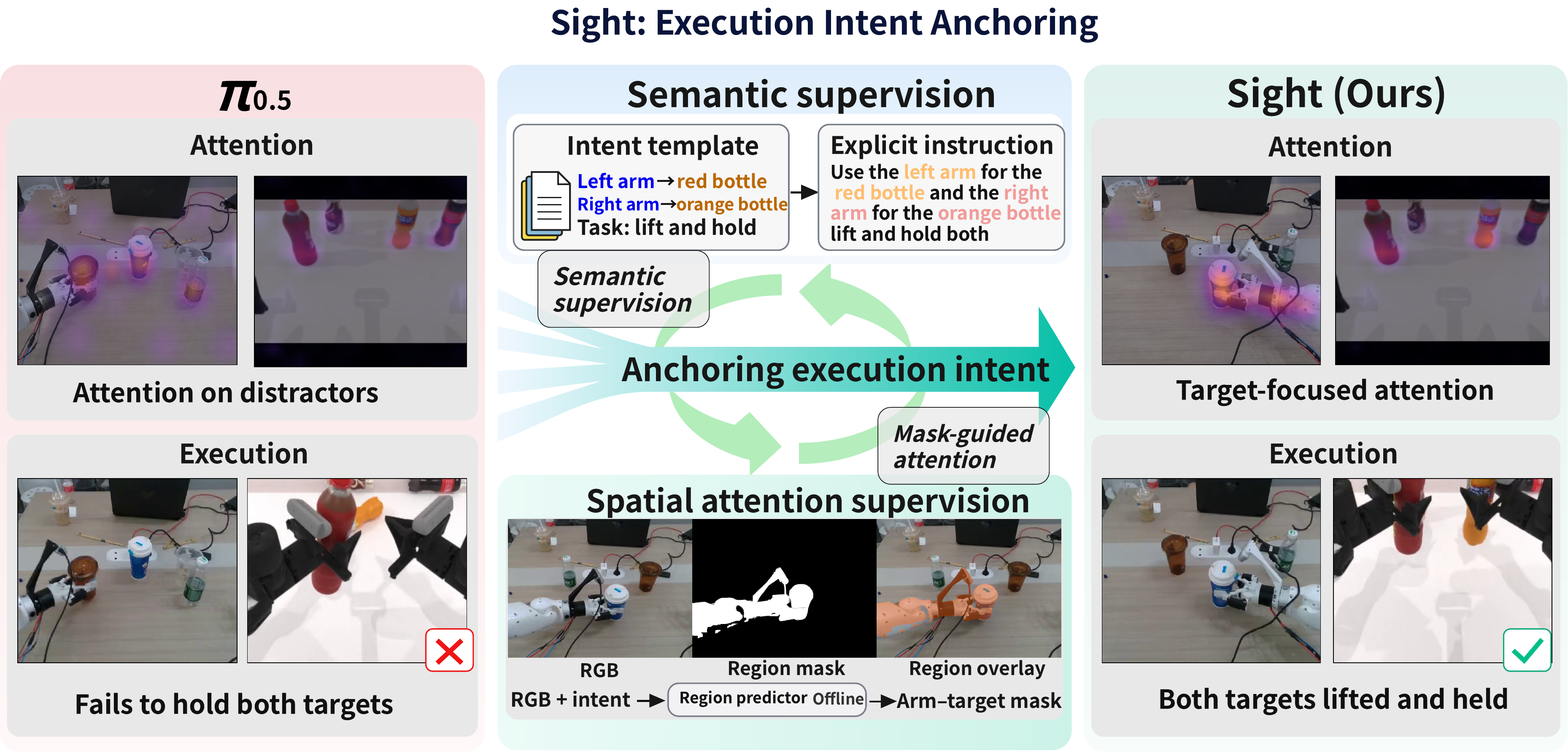}
\caption{\textbf{Sight anchors execution intent through shared semantic and spatial supervision.} The center illustrates how an explicit arm--target assignment specifies the instruction and the target-and-end-effector regions used for offline supervision. The side panels show representative attention and execution examples for the baseline and Sight. Region prediction is used during training; deployment retains the original policy interface. Quantitative task results appear in Tables~\ref{tab:main-results} and~\ref{tab:real-results}.}
\label{fig:overview}
\end{figure*}

We integrate Sight into \pifive{} post-training and evaluate it on four RoboTwin 2.0 tasks, an additional distractor condition, and the Tianji M6 robot. Sight improves mean simulation success by 13.75 percentage points (34.0\% relative) over standard post-training. Under the same explicit language, it exceeds ECR-only training by 11.50 points; the gain over ICSA-only training is 6.25 points. On OpenVLA-OFT, Sight improves mean success by 10.25 points and distractor success by 8 points. On M6, the gains are 37, 29, and 43 points for arm specification, target selection, and joint assignment, respectively.

Our contributions are threefold:
\begin{itemize}
\item We introduce \textbf{Sight} for execution-intent grounding in dual-arm manipulation. Its \textbf{ECR} component makes demonstrated task requirements and arm--target assignments explicit in the conditions used for policy learning and region prediction.
\item We develop \textbf{ICSA} to couple target-and-end-effector regions with assignment-conditioned supervision views, preserving all RGB observations and the base policy's deployment interface.
\item We evaluate the framework through simulation component and view ablations, adaptation to OpenVLA-OFT, and three categories of real-world manipulation. These comparisons quantify the complete pipeline's task gains and the contribution of spatial supervision under matched explicit language.
\end{itemize}

\section{Related Work}

\subsection{Language-Conditioned Policies and Embodied Semantics}
BC-Z and RT-1 learn task-conditioned policies; RT-2 transfers vision--language pretraining to control.~\cite{bcz,rt1,rt2} Open X-Embodiment, BridgeData V2, and DROID broaden robot training data.~\cite{oxe,bridgev2,droid} OpenVLA, Octo, and the $\pi_0$ family provide generalist policies that adapt to downstream tasks.~\cite{openvla,octo,pi0,pi05} Diffusion Policy and ACT model action sequences, with ACT, Mobile ALOHA, and ALOHA Unleashed demonstrating increasingly capable bimanual imitation.~\cite{diffusionpolicy,act,mobilealoha,alohaunleashed} These advances motivate examining how instructions select among feasible executions.

Task specification provides one interface for this selection. VIMA uses multimodal prompts, while VPP2 connects interaction semantics and motion through detailed descriptions.~\cite{vima,vpp2} ECR explicitly expresses the executing arm, target, and task requirement, and shares this assignment across policy conditioning, region labels, and auxiliary supervision views.

\subsection{Visual Grounding and Interaction-Centric Supervision}
CLIPort combines semantic and spatial pathways; PerAct predicts actions in a language-conditioned voxel representation.~\cite{cliport,peract} RVT aggregates multi-view evidence, while VoxPoser constructs language-grounded spatial value maps for planning.~\cite{rvt,voxposer} More recent approaches introduce spatial guidance into VLA learning: RoboGround uses object and placement-region masks, and ReconVLA reconstructs target regions.~\cite{roboground,reconvla} AFP predicts language-conditioned task regions and aligns policy attention during post-training, preserving the observation stream at inference.~\cite{afp}

Sight uses region prediction and attention alignment as training components, with supervision organized around a shared arm--target assignment. ECR conditions both the policy and region predictor; ICSA specifies the participating entities and supervision views. The head and active wrist supply scene-level and local evidence for single-arm operations; simultaneous dual-arm operations use all three views. Sight couples the semantic assignment to both the supervised entities and their observation sources.

\subsection{Instruction Following and Execution Consistency}
CALVIN evaluates long-horizon language-conditioned behavior, LIBERO studies knowledge transfer, and RoboCasa provides diverse household tasks.~\cite{calvin,libero,robocasa} BiGym, RoboTwin 2.0, and RoboDojo extend evaluation to bimanual manipulation and complex skills.~\cite{bigym,robotwin2,robodojo} InstructMove and \emph{Encoded but Not in Control} examine instruction dependence and the gap between represented targets and executed choices.~\cite{instructmove,groundinggap}

CAST uses counterfactual annotations, CounterAlign learns from instruction--demonstration mismatches, and CAG strengthens language-dependent action differences at inference.~\cite{cast,counteralign,cag} Gaze2Act supplies spatial intent through gaze conditioning.~\cite{gaze2act} Sight connects semantic and spatial supervision through the executing arm and designated target.

\section{Method}
\label{sec:method}

\subsection{Problem Formulation and Overview}
\label{sec:method-overview}
A visual state in dual-arm, multi-object manipulation can admit several physically feasible arm--target assignments. The instruction selects the required assignment from this set. Action imitation with an underspecified instruction can leave the executing arm and target implicit in scene--action correlations. Sight makes the demonstrated assignment explicit and connects it to the visual entities used during policy post-training.

At time $t$, the observation $o_t=(I_t^{\mathrm h},I_t^{\mathrm w_L},I_t^{\mathrm w_R},s_t)$ contains head and wrist images and the robot state. Conditioned on instruction $\ell$, the policy predicts an action chunk
\begin{equation}
 A_t\sim\pi_\theta(\cdot\mid o_t,\ell),\qquad A_t\in\mathbb R^{H\times d_a}.
 \label{eq:policy}
\end{equation}
Here $H$ is the prediction horizon and $d_a$ the action dimension. The robot executes $h_{\mathrm{exec}}\le H$ steps before observing and replanning. The instruction therefore conditions repeated decisions as the visual state changes.

For a single-active-arm operation, let $r=(e,z)$ denote the executing arm $e\in\{\mathrm L,\mathrm R\}$ and target instance $z$. The task requirement $\ell^{\mathrm{task}}$ specifies the operation, destination, or spatial relation; together with $r$, it defines the execution intent. Simultaneous dual-arm operations use a set of arm--target pairs $R$.

\begin{figure*}[!t]
\centering
\includegraphics[width=\textwidth]{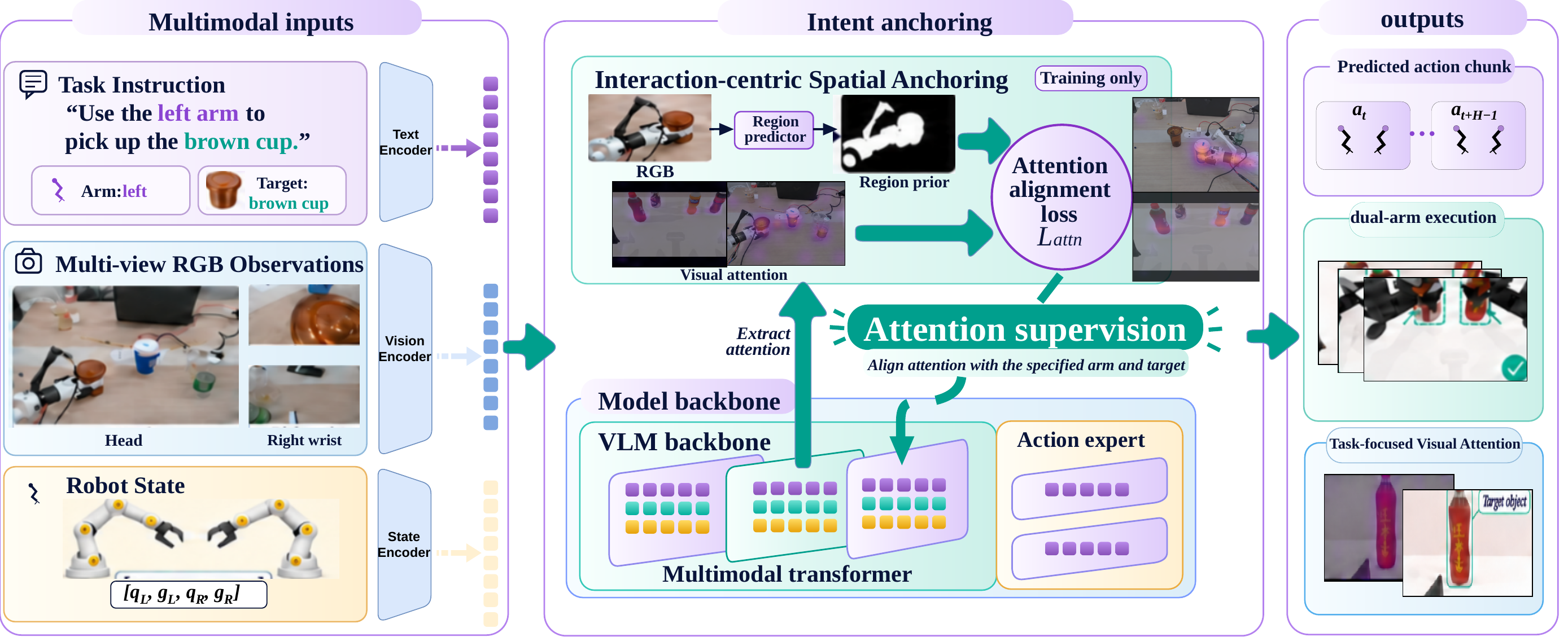}
\caption{\textbf{Sight architecture.} The instruction, multi-view RGB observations, and robot state condition the backbone and action expert. During training, offline region priors supervise attention in the selected views through $\mathcal L_{\mathrm{attn}}$. The policy receives all three RGB views and state inputs using the base \pifive{} interface. The region predictor and alignment objective are training-only components; attention overlays visualize policy attention.}
\label{fig:architecture}
\end{figure*}

Sight connects this intent to three training signals:
\begin{equation}
 (\ell^{\mathrm{task}},r)\longmapsto
 \bigl(\ell_r,\{m_{r,t}^{v}\}_{v\in\mathcal V},\{u_{r,t}^{v}\}_{v\in\mathcal V}\bigr),
 \label{eq:intent-mapping}
\end{equation}
where $\ell_r$ is the explicit instruction, $m_{r,t}^{v}$ is a regional target, and $u_{r,t}^{v}$ selects auxiliary supervision views from $\mathcal V=\{\mathrm h,\mathrm w_L,\mathrm w_R\}$. ECR constructs the language condition; ICSA constructs the spatial targets and view selection. Figure~\ref{fig:architecture} illustrates the resulting training and inference paths.

\subsection{Embodied Collaborative Representation}
\label{sec:ecr}
ECR expresses task requirements, arm allocation, and target identity in a unified language condition. A broad task description may accommodate several arm--target combinations, leaving the demonstrated choice implicit in image--action correlations. ECR makes this choice explicit so that different execution intents have distinguishable semantic conditions.

For demonstration $\tau_i$, we record the executing arm $e_i$ and target $z_i$ and construct
\begin{equation}
 \ell_i=\mathcal C\bigl(\ell_i^{\mathrm{task}},e_i,d(z_i)\bigr),
 \label{eq:language-condition}
\end{equation}
where $\mathcal C$ is a description template and $d(z_i)$ distinguishes the instance using observable attributes such as category, color, or spatial relations. The task description retains the operation and any destination or relative-position requirement. For simultaneous dual-arm demonstrations, the instruction expresses the corresponding pairings. The template binds arm names and observable target attributes to the demonstrated action.

The resulting condition is shared by action learning and region prediction. The policy learns the demonstrated action under $\ell_i$, while the region predictor uses the same instruction to locate the relevant entities. Changing the arm description changes the executing end effector and its wrist supervision; changing the target description changes the designated instance. Task requirements such as relative placement remain in the condition. ECR thus supplies the assignment that ICSA translates into spatial supervision, while preserving the operation to be learned.

\subsection{Interaction-Centric Spatial Anchoring}
\label{sec:icsa}
ICSA translates an arm--target assignment into interaction-region priors and multi-view attention supervision. It supervises both the target and active end effector and selects observation sources according to the executing arm.

\paragraph{Interaction-region priors.}
We identify the assigned instance and active end effector in keyframes using SAM~2.1 and propagate their soft regions with MatAnyone, checking instance identity in keyframes and sparse temporal inspections.~\cite{sam2,matanyone} For a single-active-arm demonstration, the label in view $v$ is
\begin{equation}
 y_{i,t}^{v}=\max\bigl(\alpha_{z_i,t}^{v},\alpha_{e_i,t}^{v}\bigr),
 \label{eq:shared-annotation}
\end{equation}
where $\alpha_z,\alpha_e\in[0,1]^{224\times224}$ are target and end-effector regions and the maximum is pixelwise. Simultaneous dual-arm operations use the union of the relevant targets and both active end effectors. Language retains the specific arm--target pairing; the union supplies the locations of the participating entities.

A language-conditioned predictor $f_\phi$ learns these regions from five-frame image clips and ECR instructions using visual features, text features, temporal aggregation, and mask, overlap, and temporal-change objectives.~\cite{afp} Before policy post-training, we freeze $f_\phi$ and cache its soft predictions offline. Average pooling produces $P=16\times16$ regional weights per view, aligned to the visual-token grid. RGB and masks share geometric augmentation; color augmentation applies only to RGB, and padded pixels have zero region weight.

\paragraph{Assignment-conditioned view supervision.}
The head view provides scene-level evidence about candidate objects and arms, while the active wrist provides local evidence tied to the executing arm. For a single active arm, we select
\begin{equation}
 u_t^v(r)=\chi_t^v\,\mathbf 1[v\in\{\mathrm h,\mathrm w(e)\}],
 \label{eq:view-selection}
\end{equation}
where $\chi_t^v$ is the image-validity flag and $\mathrm w(e)$ is the wrist camera of arm $e$. Simultaneous dual-arm operations use the union of both active wrists. All three RGB views remain inputs to the policy; this rule selects only the auxiliary supervision sources.

Changing the assigned arm changes its language description, end-effector region, and wrist supervision together. Changing the target changes the instance description and target region while retaining the view-selection rule. Supervision content and observation source therefore share an explicit assignment.

\paragraph{Region-guided attention learning.}
We extract attention from the vision--language prefix to image tokens and average over valid prefix queries and attention heads.~\cite{afp} For layer $l$,
\begin{equation}
 a_{b,j}^{(l)}=\frac{1}{N_{\mathrm{head}}|\mathcal Q_b|}
 \sum_h\sum_{q\in\mathcal Q_b}W^{(l)}_{b,h,q,j},
 \label{eq:method-attention-extraction}
\end{equation}
where $b,h,q,j$ index samples, heads, prefix queries, and image keys, and $\mathcal Q_b$ excludes padding. We supervise $L$ attention layers. We concatenate image tokens in the policy's actual order and expand the view mask to $u_{b,j}$. Attention and regional weights are normalized over the selected positions:
\begin{equation}
\begin{aligned}
 \bar a_{b,j}^{(l)}&=\frac{u_{b,j}a_{b,j}^{(l)}}{\operatorname{sg}(Z_{a,b}^{(l)})},
 &\bar m_{b,j}&=\frac{u_{b,j}m_{b,j}}{Z_{m,b}},
\end{aligned}
\label{eq:method-attention-normalization}
\end{equation}
where $Z_{a,b}^{(l)}=\max(\sum_j u_{b,j}a_{b,j}^{(l)},\epsilon)$, $Z_{m,b}=\max(\sum_j u_{b,j}m_{b,j},\epsilon)$, $\epsilon=10^{-8}$, and $\operatorname{sg}$ stops gradients. Let $\mathcal B_+$ contain samples whose selected mask mass exceeds $0.01$. We minimize
\begin{equation}
 \mathcal L_{\mathrm{attn}}=
 \frac{\sum_{l=1}^{L}\sum_{b\in\mathcal B_+}
 \|\bar a_b^{(l)}-\bar m_b\|_2^2}
 {LVP\max(1,|\mathcal B_+|)}.
 \label{eq:attention-loss}
\end{equation}
Unselected positions are zero, $V=3$, and $VP=768$ is the full image-grid length. Empty valid-sample sets yield zero auxiliary loss. This objective guides visual attention toward the assigned target and end effector while the policy learns the demonstrated actions.

\paragraph{Coupling semantic and spatial assignments.}
Language and regional supervision carry complementary information. Language specifies the operation and arm--target pairing, while the regional target identifies the participating entities. In a single-arm operation, replacing the executing arm changes both the end-effector region and the supervised wrist. In a simultaneous dual-arm operation, exchanging the pairing of the same two arms and objects leaves their regional union unchanged; the explicit language condition carries that distinction. The spatial objective therefore supervises entity localization, with relational assignment supplied by ECR. 

\subsection{Joint Policy Post-Training}
\label{sec:joint-training}
We integrate Sight into \pifive{} flow-matching post-training.~\cite{pi05,flowmatching} For normalized action chunk $A$, Gaussian noise $\varepsilon\sim\mathcal N(0,\mathrm I)$, and flow time $\eta=0.999\beta+0.001$ with $\beta\sim\operatorname{Beta}(1.5,1)$, we form
\begin{equation}
 X_\eta=(1-\eta)A+\eta\varepsilon
 \label{eq:method-flow-interpolation}
\end{equation}
and train the action expert with
\begin{equation}
 \mathcal L_{\mathrm{act}}=\mathbb E\left[
 \frac{\|v_\theta(X_\eta,\eta,o,\ell)-(\varepsilon-A)\|_F^2}{Hd_a}\right].
 \label{eq:method-action-loss}
\end{equation}
The expectation covers demonstrations, noise, and flow time. The action objective learns the demonstrated behavior, while the regional objective supervises its visual evidence under the same observation and instruction.

We preserve the action gradient and apply a one-sided projection to conflicting auxiliary gradients.~\cite{afp,pcgrad} Let $g_a=\nabla_\theta\mathcal L_{\mathrm{act}}$, $g_v=\nabla_\theta\mathcal L_{\mathrm{attn}}$, and let $M$ denote the structural support of the auxiliary gradient. With $a_M=M\odot g_a$,
\begin{equation}
\begin{aligned}
 c&=\min\left(0,\frac{\langle a_M,g_v\rangle}{\|a_M\|_2^2+\epsilon_g}\right),\\
 \widetilde g_v&=g_v-c\,a_M,\qquad g=g_a+\lambda\widetilde g_v.
\end{aligned}
\label{eq:method-gradient-projection}
\end{equation}
We use $\lambda=0.1$ and $\epsilon_g=10^{-8}$. Both gradients use the same batch and random conditions and are combined before clipping and optimization. Predictor settings and additional training details are provided in the supplementary material.

The action gradient is retained throughout the update. Projection reduces the component of auxiliary supervision that opposes it on shared parameters; parameters outside the auxiliary support receive only the action update. The objectives consequently couple demonstrated motion with region-guided visual learning under a common instruction.

\subsection{Deployment}
\label{sec:deployment}
Sight retains the base policy's action interface. At each decision, the policy receives all three RGB views, the robot state, and the task instruction and generates an action chunk. The region predictor, cached masks, and auxiliary objective are training-only components. Changing the requested arm or target changes the language condition supplied to the policy, while subsequent decisions use the updated observations and the current task requirement.

Complete observations retain the context needed to interpret the task, including reference objects for relative placement and neighboring objects around an approach. Auxiliary view selection affects attention supervision during training, while closed-loop execution uses all available RGB views and the current instruction. 

\section{Experiments}
\label{sec:experiments}
We evaluate the task gains of the complete Sight pipeline, the contribution of spatial supervision under explicit language, and the effect of auxiliary supervision views. Four RoboTwin 2.0 tasks and an additional distractor condition measure manipulation performance. The Tianji M6 evaluation covers arm specification, multi-candidate target selection, and joint arm--target assignment.

\paragraph{Comparison logic.}
The component and view comparisons isolate different parts of the training design. ECR-only and Sight share explicit instructions, so their comparison measures the contribution of region-guided attention learning. ICSA-only uses the original language throughout region prediction and policy learning; its comparison with Sight measures the complete semantic-conditioning change. Sight-AllViews retains the explicit language and regional targets while changing which views enter the auxiliary objective. 
\paragraph{Qualitative execution sequences.}
Figure~\ref{fig:rollouts} pairs policy attention with simulation and M6 execution sequences. Under the same left-arm, blue-cup instruction, ten chronological frames contrast baseline failure and Sight success. Additional Sight sequences illustrate executions under other arm--target assignments.

\twocolumn[{%
\begin{minipage}{\textwidth}
\centering
\includegraphics[width=\textwidth,height=0.48\textheight,keepaspectratio]{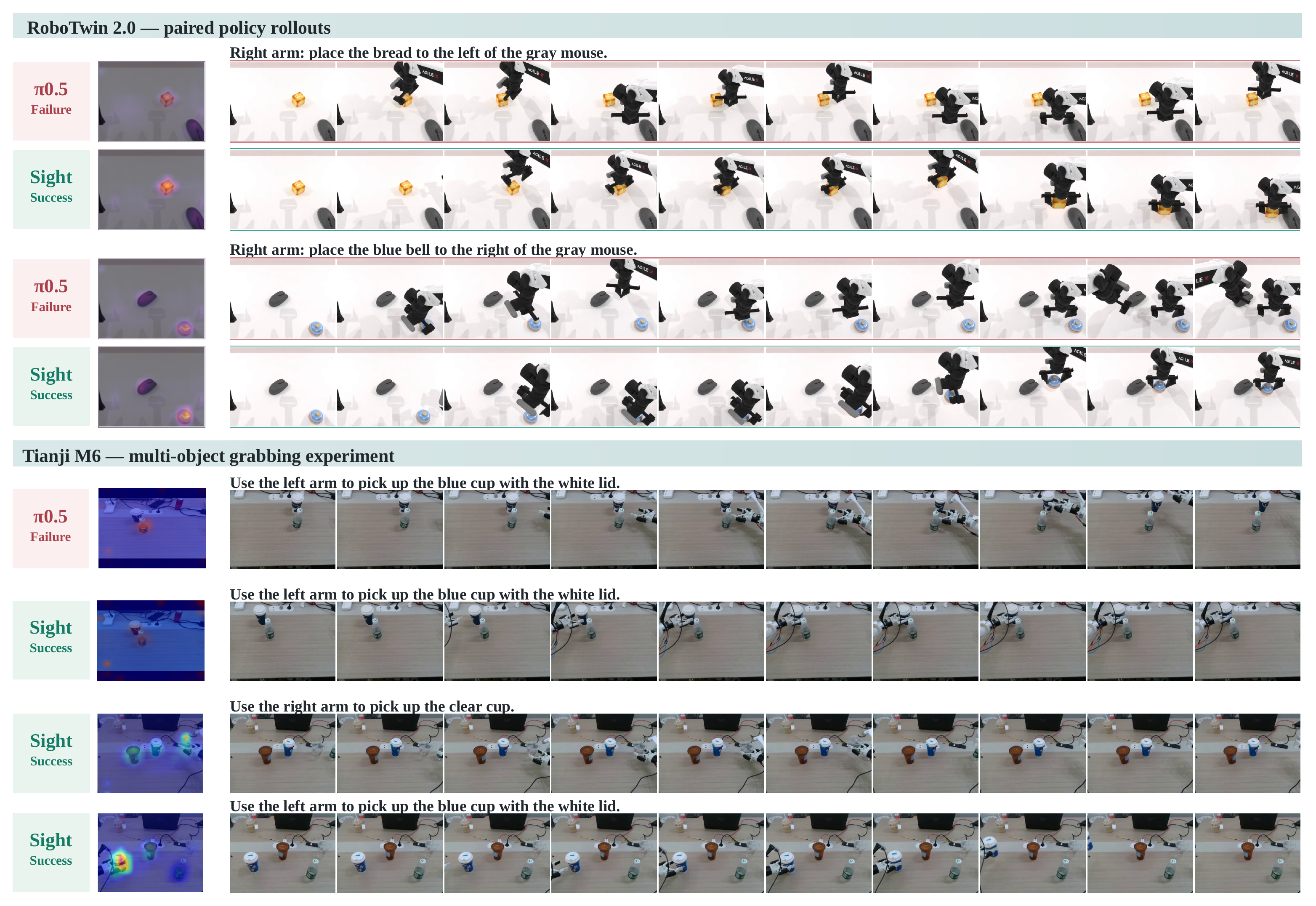}
\captionof{figure}{Policy attention and instruction-conditioned execution. Each row pairs an attention visualization (left) with ten chronological execution frames (right). Top: relation-conditioned placement in RoboTwin 2.0. Bottom: M6 manipulation. Under the same left-arm, blue-cup instruction, the baseline fails to lift the cup, whereas Sight uses the requested arm and lifts it. The final two rows show additional Sight executions with specified arms and targets. Real-world trials have different initial layouts.}
\label{fig:rollouts}
\par\vspace{\textfloatsep}
\end{minipage}
}]

\subsection{Experimental Setup}
\paragraph{Tasks and training.}
We select coaster placement, dual-arm holding, left-relation placement, and right-relation placement from RoboTwin 2.0.~\cite{robotwin2} These tasks cover object manipulation, simultaneous dual-arm execution, and placement relative to a reference object. We use 50 demonstrations per task, splitting the 200 complete episodes into 160 training and 40 validation episodes. All configurations start from $\pi_{0.5}^{\mathrm{base}}$ and use $224\times224$ images, a 64-step action horizon, a global batch size of 16, and the EMA checkpoint at step 19,000. Data splits, action statistics, optimizer settings, and learning-rate schedules are shared; the recorded training recipe uses seed 42. Each task uses the same 100 predefined test entries. Additional implementation details are provided in the supplementary material.

\paragraph{Component configurations.}
We compare standard \pifive{} post-training, \pifive{} + ECR, \pifive{} + ICSA, and full Sight. ECR and Sight use the same explicit language conditions, so their comparison measures the increment from spatial anchoring. The ICSA-only predictor and masks are independently trained and cached using the original language. Its comparison with Sight measures the full semantic-conditioning pipeline across region prediction, policy training, and inference. All configurations receive head and both wrist RGB views and retain the base action interface.

\paragraph{Success criteria.}
Simulation results use the environment task-success criterion. The distractor condition retains the original two-target holding criterion. We report the proportion of successful trials and average task success rates with equal weights. Real-world success requires completing the operation while satisfying the corresponding arm and target specification, as detailed in Sec.~\ref{sec:real-world}.

\subsection{Manipulation Performance and Component Analysis}
Table~\ref{tab:main-results} reports the four manipulation tasks and the distractor condition.

\begin{table*}[!t]
\centering\small
\caption{RoboTwin 2.0 success rates (\%). Each task uses 100 test entries; the mean weights the four tasks equally. The distractor test adds two non-target bottles to dual-arm holding without additional demonstrations. Bold marks column maxima.}
\label{tab:main-results}
\begin{tabular}{@{}lrrrrrr@{}}\toprule
Method & \shortstack{Coaster\\placement} & \shortstack{Dual-arm\\holding} & \shortstack{Left-relation\\placement} & \shortstack{Right-relation\\placement} & Mean & \shortstack{Two additional\\distractors} \\\midrule
\pifive{} & 62 & 42 & 31 & 27 & 40.50 & 8 \\
\pifive{} + ECR & 64 & 43 & 33 & 31 & 42.75 & 7 \\
\pifive{} + ICSA & 70 & 55 & \textbf{41} & 26 & 48.00 & 12 \\
\textbf{Sight} & \textbf{85} & \textbf{56} & \textbf{41} & \textbf{35} & \textbf{54.25} & \textbf{16} \\\bottomrule
\end{tabular}
\end{table*}

\paragraph{Manipulation performance.}
Sight achieves 54.25\% mean success, compared with 40.50\% for standard \pifive{} post-training: a gain of 13.75 percentage points, or 34.0\% relative improvement. Gains are 23 points for coaster placement, 14 for dual-arm holding, and 10 and 8 for left- and right-relation placement. Improvements therefore extend across object manipulation, dual-arm execution, and relation-conditioned placement.

\paragraph{Semantic and spatial components.}
The ECR-only and ICSA-only configurations reach mean success rates of 42.75\% and 48.00\%, respectively. Full Sight exceeds them by 11.50 and 6.25 points, respectively. It improves on ECR-only training in all four tasks and on ICSA-only training in three tasks, matching its 41\% success on left-relation placement. The 11.50-point gain over ECR-only training isolates the contribution of spatial supervision under matched explicit language. Combining the semantic and spatial components gives the highest mean success across the four tasks.

\paragraph{Additional candidate objects.}
We add two non-target bottles to the dual-arm holding scene while retaining the two targets and the original success criterion. Policies are evaluated on 100 fixed scenes without additional demonstrations. Sight reaches 16\% success, compared with 8\% for the baseline, 7\% for ECR alone, and 12\% for ICSA alone. The eight-point gain over standard post-training measures performance under this additional-candidate condition, which requires completing the original operation amid competing objects.

\subsection{Execution-Conditioned View Supervision}
\label{sec:ablation}
We compare Sight with Sight-AllViews, which uses the same explicit instructions, predictor, region definitions, RGB inputs, and 19,000-step budget but replaces Eq.~\eqref{eq:view-selection} with $u_t^v=\chi_t^v$. All valid views then contribute to attention alignment, changing its normalization support accordingly. During single-active-arm operations, this adds the inactive wrist; simultaneous dual-arm operations use all three views in both configurations. Of the 200 demonstrations, 150 have a single active arm and 50 use both arms.

\begin{table}[!t]
\centering\small
\caption{Auxiliary-view ablation (success, \%). Both policies receive three RGB views. For single-arm supervision, Sight selects the head and active wrist; AllViews uses every valid view. Both supervise all three views in dual-arm tasks.}
\label{tab:ablation}
\begin{tabular}{@{}lrr@{}}\toprule
Task & Sight & AllViews \\\midrule
Coaster placement & \textbf{85} & 81 \\
Dual-arm holding & \textbf{56} & 49 \\
Left-relation placement & \textbf{41} & 39 \\
Right-relation placement & \textbf{35} & 29 \\\midrule
Mean & \textbf{54.25} & 49.50 \\
Additional distractors & \textbf{16} & 12 \\\bottomrule
\end{tabular}
\end{table}

\begin{table}[!t]
\centering\small
\caption{OpenVLA-OFT adaptation (success, \%). The two configurations share data, action settings, training budgets, and test scenes. The mean weights the four tasks equally.}
\label{tab:cross-policy}
\begin{tabular}{@{}lcc@{}}\toprule
Task & \shortstack{OpenVLA\\-OFT} & \shortstack{OpenVLA-OFT\\+ Sight} \\\midrule
Coaster placement & 20 & \textbf{53} \\
Dual-arm holding & 9 & \textbf{10} \\
Left-relation placement & 5 & \textbf{12} \\
Right-relation placement & \textbf{11} & \textbf{11} \\\midrule
Mean & 11.25 & \textbf{21.50} \\
Additional distractors & 0 & \textbf{8} \\\bottomrule
\end{tabular}
\end{table}

Table~\ref{tab:ablation} reports a 4.75-point mean gain over AllViews, with per-task gains of 4, 7, 2, and 6 points; distractor success increases by 4 points. The view-selection rule differs on the 150 single-active-arm demonstrations. Both configurations use all three supervision views on dual-arm samples, so the holding result measures downstream performance under the two overall training configurations. Across the four tasks, assignment-conditioned supervision yields higher mean success with the same RGB inputs and explicit language.

\subsection{Adapting Sight to OpenVLA-OFT}
\label{sec:cross-policy}
We apply Sight to OpenVLA-OFT~\cite{openvlaoft}, retaining OFT action learning with a 25-step horizon. Both configurations share initialization, 160 training and 40 validation episodes, three RGB views, action definitions, test scenes, and training budgets. Mean success increases from 11.25\% to 21.50\%, a gain of 10.25 points (Table~\ref{tab:cross-policy}). Gains are 33, 1, and 7 points for coaster placement, dual-arm holding, and left-relation placement; right-relation placement remains at 11\%. Distractor success increases from 0\% to 8\%.

\twocolumn[{%
\begin{minipage}{\textwidth}
\centering
\includegraphics[width=\textwidth,height=0.70\textheight,keepaspectratio]{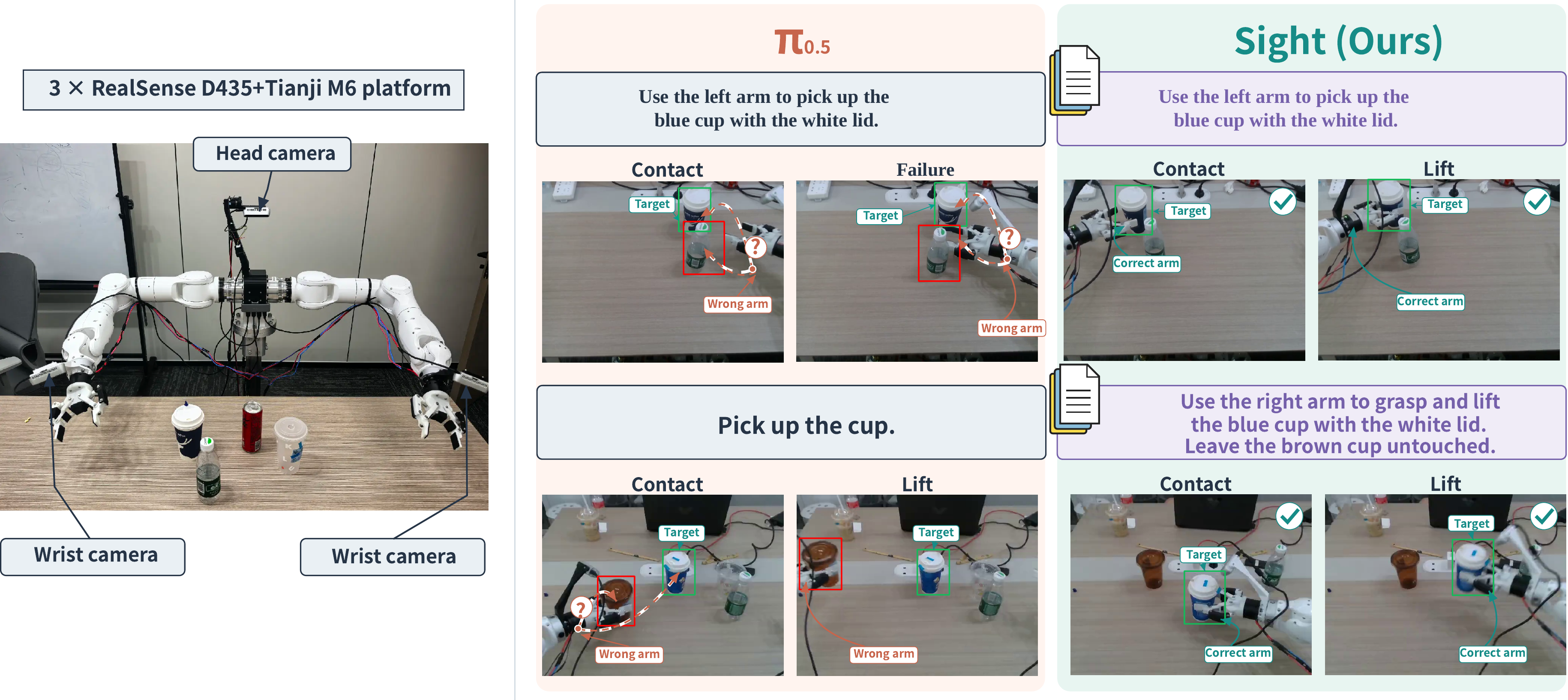}
\captionof{figure}{M6 platform and prompt-conditioned execution. Left: head and wrist cameras. Upper right: under the same left-arm, blue-cup instruction, the baseline uses the other arm and approaches competing objects; Sight lifts the specified cup. Lower right: the baseline lifts a brown cup under a broad prompt; Sight lifts the blue cup with the specified right arm under an explicit prompt. Both policy and prompt vary in this lower comparison; annotations refer to the explicit assignment.}
\label{fig:real-platform}
\par\vspace{\textfloatsep}
\end{minipage}
}]

\subsection{Real-World Manipulation and Execution Intent}
\label{sec:real-world}
We evaluate three execution-intent categories on M6 (Table~\ref{tab:real-results}); Figures~\ref{fig:rollouts} and~\ref{fig:real-platform} show representative executions.

\begin{table}[H]
\centering\small
\setlength{\tabcolsep}{4pt}
\caption{Real-world execution-intent tasks on M6 (success, \%). The mean weights the three task categories equally and is rounded to the nearest integer. Figure~\ref{fig:real-platform} provides representative examples.}
\label{tab:real-results}
\begin{tabular}{@{}lcc@{}}\toprule
Task category & \pifive{} & \shortstack{\pifive{}\\+ Sight} \\\midrule
Arm specification & 62 & \textbf{99} \\
Multi-candidate target selection & 47 & \textbf{76} \\
Joint arm--target assignment & 31 & \textbf{74} \\\midrule
Mean success rate & 47 & \textbf{83} \\\bottomrule
\end{tabular}
\end{table}

\paragraph{Platform and data.}
The Tianji M6 uses head and left/right wrist RealSense D435+ cameras. We collect 100 teleoperated trajectories (18,315 frames at 30\,Hz) involving brown, blue, and transparent cups and a green-labeled bottle, splitting them into 90 training and 10 validation episodes. Training uses 20,000 updates; 16-dimensional states and actions are padded to 32 dimensions.

\paragraph{Task categories and instruction switching.}
\emph{Arm specification} fixes the arm for a known target; \emph{multi-candidate target selection} specifies the target; \emph{joint arm--target assignment} specifies both. For joint assignment, we test counterfactual instruction following in the \textbf{same scene under different instructions}: object identities and placements remain fixed across trials while the instructed arm--target pairing changes. Success requires completing the operation with the specified pairing.

\paragraph{Execution examples.}
Figure~\ref{fig:real-platform} complements this protocol: the upper comparison shows baseline failure and Sight success under a shared explicit instruction; the lower comparison illustrates baseline execution under a broad prompt and Sight execution under an explicit arm--target prompt.

\paragraph{Results.}
Sight improves success by 37, 29, and 43 percentage points for arm specification, target selection, and joint assignment, respectively (Table~\ref{tab:real-results}). Rounded mean success increases from 47\% to 83\%.
\FloatBarrier

\section{Conclusion}
Sight couples ECR language conditions with ICSA regional and view supervision through shared arm--target assignments, retaining complete RGB inputs and the original deployment interface. On four RoboTwin 2.0 tasks, mean success improves by 13.75 points over standard \pifive{} post-training and by 11.50 points over ECR-only training under the same explicit language. On OpenVLA-OFT, mean simulation success improves by 10.25 points. On Tianji M6, rounded mean success across arm specification, target selection, and joint assignment increases from 47\% to 83\%. The results support combining explicit execution assignments with region-guided policy learning.

\clearpage
\section*{LLM Usage Statement}
OpenAI Codex assisted with literature organization, manuscript drafting,
translation, language editing, preliminary figure-layout exploration, and
LaTeX preparation. The authors are responsible for verifying scientific claims, bibliographic
information, experimental results, and the final manuscript.
{\small
\bibliographystyle{ieeenat_fullname}
\bibliography{main}
}
\clearpage
\maketitlesupplementary
\appendix
\section{Training Details}\label{sec:training-details}

\subsection{Attention Extraction and Normalization}
Let $W^{(l)}_{b,h,q,j}$ denote attention at layer $l$, with sample, head, prefix-query, and image-key indices $b,h,q,j$. The set $\mathcal Q_b$ contains non-padding vision--language prefix positions. We extract image attention as
\begin{equation}
 a^{(l)}_{b,j}=\frac{1}{N_{\mathrm{head}}|\mathcal Q_b|}
 \sum_h\sum_{q\in\mathcal Q_b}W^{(l)}_{b,h,q,j}.
 \label{eq:attention-extraction}
\end{equation}
These weights are the image-key portion of prefix attention, before renormalization over images. Expanding the view-selection rule into token mask $u_{b,j}$ and aligning the predicted soft mask $m_{b,j}$ to the same token order gives
\begin{equation}
\begin{aligned}
 Z^{(l)}_{a,b}&=\max\left(\sum_j u_{b,j}a^{(l)}_{b,j},\epsilon\right),\\
 Z_{m,b}&=\max\left(\sum_j u_{b,j}m_{b,j},\epsilon\right),
\end{aligned}
\end{equation}
\begin{equation}
 \bar a^{(l)}_{b,j}=\frac{u_{b,j}a^{(l)}_{b,j}}{\operatorname{sg}(Z^{(l)}_{a,b})},
 \qquad \bar m_{b,j}=\frac{u_{b,j}m_{b,j}}{Z_{m,b}}.
 \label{eq:attention-normalization}
\end{equation}
We use stop-gradient $\operatorname{sg}$ on the attention normalizer and set $\epsilon=10^{-8}$. Unselected positions have zero weight. The frozen predictor supplies fixed mask targets. The attention loss uses samples with sufficient selected mask mass,
\begin{equation}
 \mathcal B_+=\left\{b:\sum_{j=1}^{VP}u_{b,j}m_{b,j}>0.01\right\}.
\end{equation}
We align pooled regional targets with image-token attention and scale the loss by the full grid length $VP=768$. For single-arm operations, Sight supervises up to 512 positions and Sight-AllViews supervises up to 768. Both supervise up to 768 positions for simultaneous dual-arm operations.

\subsection{Action Objective}
For normalized action chunk $A\in\mathbb R^{H\times d_a}$ and noise $\varepsilon\sim\mathcal N(0,\mathrm I)$, we sample $\beta\sim\operatorname{Beta}(1.5,1)$, set $\eta=0.999\beta+0.001$, and form
\begin{equation}
 X_\eta=(1-\eta)A+\eta\varepsilon.
\end{equation}
The action objective is
\begin{equation}
 \mathcal L_{\mathrm{act}}=\mathbb E\left[
 \frac{\|v_\theta(X_\eta,\eta,o,\ell)-(\varepsilon-A)\|_F^2}{Hd_a}\right].
 \label{eq:action-loss}
\end{equation}
The expectation averages over demonstrations, noise, and flow time.

\subsection{Auxiliary Gradient Projection}
Let $g_a=\nabla_\theta\mathcal L_{\mathrm{act}}$ and $g_v=\nabla_\theta\mathcal L_{\mathrm{attn}}$ be flattened into the same parameter space, with zeros outside the auxiliary path. The mask $M$ denotes the structural support of that path, including the relevant layers and packed K/V subparameters, and satisfies $M\odot g_v=g_v$. This support is fixed by the computation graph. With $a_M=M\odot g_a$,
\begin{equation}
\begin{aligned}
 c&=\min\left(0,\frac{\langle a_M,g_v\rangle}{\|a_M\|_2^2+\epsilon_g}\right),\\
 \widetilde g_v&=g_v-c\,a_M,\qquad g=g_a+\lambda\widetilde g_v.
\end{aligned}
\label{eq:gradient-projection}
\end{equation}
We set $\lambda=0.1$ and $\epsilon_g=10^{-8}$. The projection reduces the component of the auxiliary gradient that opposes the action gradient on shared parameters. Both objectives use the same inputs and random draws. We clip the combined gradient before the AdamW update.

\subsection{Region Predictor Training}
Our text-conditioned region predictor combines a pretrained MobileNetV3-Large visual backbone with a frozen CLIP ViT-B/32 text encoder. The deformable transformer uses three feature levels, one encoder layer, one decoder layer, one region query, and a hidden dimension of 256. A feature-wise affine adapter conditions the decoder states on the episode instruction before mask prediction. The decoder combines queries over time using learned weights, and the mask head aggregates feature-pyramid features over time. The training target is the soft union of the assigned target and active end-effector regions. We weight sigmoid focal loss, Dice loss, and the temporal loss by $5$, $1$, and $1$, respectively. The temporal term is the mean squared error between adjacent-frame changes in the predicted and target masks.

\paragraph{Clip sampling.}
We export contiguous runs of at least five valid frames within each episode and camera stream. The predictor uses head and active-wrist runs for single-arm episodes and all three views for simultaneous dual-arm episodes. The loader places a clip anchor every five frames within each run. For training, it scales the five offsets by a factor drawn uniformly from $\{0.5,1,2,3,4,5\}$, rounds them down, adds a random integer shift from $0$ to $4$, and reverses their order with probability $0.5$. Frame indices wrap within the same run. We apply photometric and motion transforms, resizing, and cropping to the training clips. M6 retains the default horizontal flip with probability $0.5$. The RoboTwin runs disable this transform to preserve left/right instruction semantics. RoboTwin validation uses consecutive, unjittered offsets.

\paragraph{Optimization and checkpoint selection.}
The predictor uses AdamW for 100 epochs with seed 42, weight decay $10^{-4}$, and gradient clipping at $0.1$. The initial learning rate is $2\times10^{-4}$ for the main group and $2\times10^{-5}$ for the visual backbone and deformable reference-point and sampling-offset projections. The rates drop by a factor of ten after epochs 18 and 30, giving main-group rates of $2\times10^{-5}$ and $2\times10^{-6}$. We use a global batch of 48 clips. The explicit-language RoboTwin predictor uses 8 GPUs with 6 clips per GPU. The original-language predictor uses 4 GPUs with 12 clips per GPU. Both fit on the same 160 training episodes and select the checkpoint with the lowest loss on the 40 held-out validation episodes. The selected epochs are 3 and 2, respectively, counted from one. For M6, the predictor fits on the 90 training episodes with 4 GPUs and 12 clips per GPU. The cache uses the final epoch-100 checkpoint. The 10 validation episodes are excluded from predictor fitting.

\subsection{Offline Region Cache}
We freeze the selected predictor and process each episode and camera separately. We resize RGB frames to $224\times224$ with aspect-ratio-preserving padding, apply the episode instruction to consecutive non-overlapping chunks of five frames, and keep the actual length of the last chunk. The predictor interpolates mask logits to the input grid before applying the sigmoid. We set the padded area to zero and store $\operatorname{round}(255\alpha)$ as an 8-bit array indexed by episode, camera, and frame. RoboTwin uses per-camera HDF5 files, while M6 uses grayscale PNGs. At policy-training time, the loader divides the stored values by 255 and applies the frame-validity flags. The same geometric augmentation is applied to each RGB--mask pair. We average each $14\times14$ block of the aligned mask to obtain the $16\times16$ image-token target grid.

\paragraph{Cache variants.}
The RoboTwin Sight cache contains 65,014 camera frames across the selected views. ICSA-only fits a separate predictor with the original task instructions and uses those same instructions to regenerate predictions for the corresponding views. Sight-AllViews reads the unchanged Sight cache on shared views and adds 23,417 inactive-wrist predictions for the 150 single-arm episodes. The additional predictions use the same frozen Sight checkpoint, RGB preprocessing, and episode instructions. The added views are gated by RGB validity, with padded pixels set to zero. The shared views retain their original audited validity flags. Both configurations already include all three views for simultaneous dual-arm episodes. The M6 cache contains 36,630 camera frames from the head and active wrist.

\section{Data and Comparison Groups}\label{sec:data-groups}
\paragraph{Configurations.}
We evaluate standard action post-training, ECR-only, ICSA-only, Sight, and Sight-AllViews. The standard and ICSA-only policies use the original RoboTwin 2.0 instructions. ECR-only, Sight, and Sight-AllViews use the same explicit ECR instructions.

\paragraph{Optimization.}
The policy learning rate warms up for 100 steps to $5\times10^{-5}$, decays to $5\times10^{-6}$ by step 12,000, and remains fixed thereafter. AdamW uses $(\beta_1,\beta_2)=(0.9,0.95)$, $\epsilon_{\mathrm{opt}}=10^{-8}$, and weight decay $10^{-10}$. Gradient clipping is 1.0 and EMA decay is 0.99. Inputs, horizon, batch size, and update budgets follow the main text.

\section{Evaluation Protocol and Metrics}
\label{sec:evaluation-statistics}
\paragraph{Simulation.}
We evaluate coaster placement, dual-arm holding, left-relation placement, and right-relation placement on RoboTwin 2.0. Each condition uses 100 predefined test entries and the environment's task-success criterion. Entries rejected by the expert gate remain in the denominator and count as failures. We average the four task success rates with equal weight. A separate condition adds two non-target bottles at predefined positions to the dual-arm holding scene and uses the same holding criterion.

\begin{samepage}
\paragraph{Real-world evaluation.}
We compare \pifive{} and \pifive{} + Sight on arm specification, target selection, and joint assignment. For trial $i$, $S_i$ indicates task completion and $C_i$ indicates compliance with the specified assignment. We define $Y_i=S_i C_i$ and compute
\begin{equation}
 \mathrm{SR}=\frac{1}{N}\sum_{i=1}^{N}S_i,\qquad
 \mathrm{ISR}=\frac{1}{N}\sum_{i=1}^{N}Y_i.
 \label{eq:success-metrics}
\end{equation}
\end{samepage}
Here $N$ counts all started trials. Timeouts, trials with no action, trials requiring manual intervention, and trials with unassessable compliance stay in the denominator and contribute zero to ISR. The main real-world table uses this assignment-aware success criterion. Its mean weights the three categories equally and rounds to the nearest integer.

\FloatBarrier
\section{Comparisons with AFP and ReconVLA}
\label{sec:external-method-comparisons}
We compare the complete post-training pipelines on the four RoboTwin 2.0 tasks and the additional distractor condition. AFP and ReconVLA use the same split of the official \textit{demo\_clean} demonstrations, with 40 training and 10 validation trajectories per task (160/40 total). Both use 19,000 policy updates, a global batch size of 16, and training seed 42. The comparisons use the step-19,000 EMA weights.

\subsection{Comparison with AFP}
\label{sec:comparison-afp}
We compare a multi-view adaptation of AFP with Sight on the same \pifive{} base policy. AFP uses the official RoboTwin 2.0 instructions for training and evaluation. Its auxiliary loss supervises all three decoded views using an original-language region predictor. Sight uses explicit ECR instructions that specify the executing arm and target alongside the task goal. The same arm--target assignment also determines Sight's regional targets and supervision views. We match the 100 test seeds per clean task and reuse the same 100 frozen layouts for the two-distractor condition.

Table~\ref{tab:comparison-afp} compares the complete pipelines under their respective instruction formats. Sight averages 54.25\% across the four clean tasks and AFP averages 47.00\%. Sight gains 28 percentage points on dual-arm holding. AFP leads by 8 points on left-relation placement. Both reach 35\% on right-relation placement. Under distractors, Sight succeeds in 16 of 100 scenes and AFP in 6 of 100. There are 11 Sight-only successes and one AFP-only success.

\begin{table}[H]
\centering\small
\caption{\textbf{AFP and Sight on RoboTwin 2.0.} Both use \pifive{}. Task success rates and their four-task mean are in percent. $\Delta$ is the Sight--AFP difference in percentage points.}
\label{tab:comparison-afp}
\begin{tabular}{@{}lrrr@{}}
\toprule
Evaluation condition & AFP & Sight & $\Delta$ \\
\midrule
Coaster placement        & 76    & 85    & $+9$ \\
Dual-arm holding         & 28    & 56    & $+28$ \\
Left-relation placement  & 49    & 41    & $-8$ \\
Right-relation placement & 35    & 35    & $0$ \\
Four-task mean           & 47.00 & 54.25 & $+7.25$ \\
\midrule
Two additional distractors & 6 & 16 & $+10$ \\
\bottomrule
\end{tabular}
\end{table}

\subsection{Comparison with ReconVLA}
\label{sec:comparison-reconvla}
We initialize ReconVLA from the authors' pretrain-checkpoint-10388 and fine-tune it on RoboTwin 2.0 with the official instructions. Training updates the Qwen2-7B backbone, SigLIP vision encoder, and reconstruction branch. The FLUX VAE stays frozen. Training uses four A800 GPUs with fully sharded data parallelism. The model uses action-token cross-entropy and reconstructs the region cropped from the head-camera image using the target mask.

We tile the head and two wrist RGB views into a $384\times384$ image and supply a 14-dimensional robot state. We adapt the action output from $5\times7$ to $64\times14$, represented as 896 autoregressive tokens. Twelve joint dimensions encode relative commands, and two gripper dimensions encode absolute commands. Table~\ref{tab:comparison-reconvla} compares this RoboTwin 2.0 adaptation with Sight.

\begin{table}[H]
\centering\small
\caption{\textbf{ReconVLA and Sight on RoboTwin 2.0.} Task success rates and their four-task mean are in percent. $\Delta$ is the Sight--ReconVLA difference in percentage points.}
\label{tab:comparison-reconvla}
\begin{tabular}{@{}lrrr@{}}
\toprule
Evaluation condition & ReconVLA & Sight & $\Delta$ \\
\midrule
Coaster placement        & 1    & 85    & $+84$ \\
Dual-arm holding         & 0    & 56    & $+56$ \\
Left-relation placement  & 1    & 41    & $+40$ \\
Right-relation placement & 1    & 35    & $+34$ \\
Four-task mean           & 0.75 & 54.25 & $+53.50$ \\
\midrule
Two additional distractors & 0 & 16 & $+16$ \\
\bottomrule
\end{tabular}
\end{table}

\FloatBarrier
\section{Arm Specification and Target Selection on M6}
\label{sec:m6-instruction-switching}
Table~\ref{tab:m6-instruction-switching} presents the arm-specification and multi-candidate target-selection results from the main paper with example command pairs. The first pair varies the requested arm for a fixed target. The second varies the requested target for a fixed arm. We score each trial using the assignment-aware criterion in Eq.~\eqref{eq:success-metrics}. Sight improves success by 37 percentage points for arm specification and 29 points for target selection.

\begin{table}[H]
\centering\small
\caption{\textbf{Arm specification and target selection on M6.} Successful trials satisfy the requested assignment and complete the task.}
\label{tab:m6-instruction-switching}
\begin{tabular*}{\columnwidth}{@{\extracolsep{\fill}}lrr@{}}
\toprule
& \multicolumn{2}{c}{\shortstack{Successful\\trials (\%)}} \\
\cmidrule(l){2-3}
Example command pair & \pifive{} & Sight \\
\midrule
\begin{tabular}[c]{@{}l@{}}\textbf{Switch arm}\\Left/right arm $\to$ blue cup\end{tabular}
  & 62 & 99 \\
\addlinespace
\begin{tabular}[c]{@{}l@{}}\textbf{Switch target}\\Left arm $\to$ blue/brown cup\end{tabular}
  & 47 & 76 \\
\bottomrule
\end{tabular*}
\end{table}

\FloatBarrier

\end{document}